%% file: root.tex
\documentclass[letterpaper, 10 pt, conference]{ieeeconf}  

\IEEEoverridecommandlockouts                              

\usepackage{cite}
\usepackage{amsmath,amssymb,amsfonts}
\usepackage{algorithmic}
\usepackage{graphicx}
\usepackage{textcomp}
\usepackage{xcolor}
\usepackage{booktabs}
\usepackage{graphicx}
\usepackage[colorlinks=true, allcolors=blue]{hyperref}
\usepackage{censor}

\newif\ifblind
\blindtrue 

\ifblind
    \protect\RestartCensoring
\else
    \protect\StopCensoring
\fi

\title{\LARGE \bf
VinePT-Map: Pole-Trunk Semantic Mapping for Resilient Autonomous Robotics in Vineyards
}

\author{
Giorgio Audrito$^{1, 2}$, Mauro Martini$^{1, 3}$, Alessandro Navone$^{1, 3}$, Giorgia Galluzzo$^{1}$ and Marcello Chiaberge$^{1, 3}$  
\thanks{This work was supported by the PoliTO Interdepartmental Centre for Service Robotics (PIC4SeR).} 
\thanks{$^{1}$Department of Electronics and Telecommunications, Politecnico di Torino, 10129, Torino, Italy.}
\thanks{\tt\footnotesize $^{2}$ giorgio\_audrito@polito.it}%
\thanks{\tt\footnotesize $^{3}$ \{name\}.\{surname\}@polito.it}
}

\begin{document}

\maketitle

\begin{abstract}
Reliable long-term deployment of autonomous robots in agricultural environments remains challenging due to perceptual aliasing, seasonal variability, and the dynamic nature of crop canopies. Vineyards, characterized by repetitive row structures and significant visual changes across phenological stages, represent a pivotal field challenge, limiting the robustness of conventional feature-based localization and mapping approaches. This paper introduces VinePT-Map, a semantic mapping framework that leverages vine trunks and support poles as persistent structural landmarks to enable season-agnostic and resilient robot localization. The proposed method formulates the mapping problem as a factor graph, integrating GPS, IMU, and RGB-D observations through robust geometrical constraints that exploit vineyard structure. An efficient perception pipeline based on instance segmentation and tracking, combined with a clustering filter for outlier rejection and pose refinement, enables accurate landmark detection using low-cost sensors and onboard computation. To validate the pipeline, we present a multi-season dataset for trunk and pole segmentation and tracking. Extensive field experiments conducted across diverse seasons demonstrate the robustness and accuracy of the proposed approach, highlighting its suitability for long-term autonomous operation in agricultural environments.
\end{abstract}

\section{Introduction}
The deployment of autonomous robotic systems in agriculture represents a pivotal step toward addressing global labor shortages and increasing operational efficiency \cite{zhai2020decision}. Although considerable progress has been made in recent years \cite{cerrato2024deep, martini2023enhancing}, achieving a reliable and persistent deployment in real-world agricultural contexts remains a formidable challenge. Most prior work has concentrated on row-based crops \cite{Bigelow:263079}, and has tackled challenges including localization \cite{winterhalter2021localization, diao2025localization}, path planning \cite{salvetti2023waypoint}, navigation \cite{wu2025review}, harvesting \cite{hua2025harvesting, droukas2023survey}, and vegetation assessment \cite{feng2020yield}. The dynamic, visually repetitive, and constantly changing nature of agricultural environments poses significant challenges to perception, localization, and mapping, making robot deployment far from straightforward \cite{SLAMAgri} \cite{AgriRobot}. 

\begin{figure}[ht]
    \centering
    \includegraphics[width=\columnwidth, angle = 0]{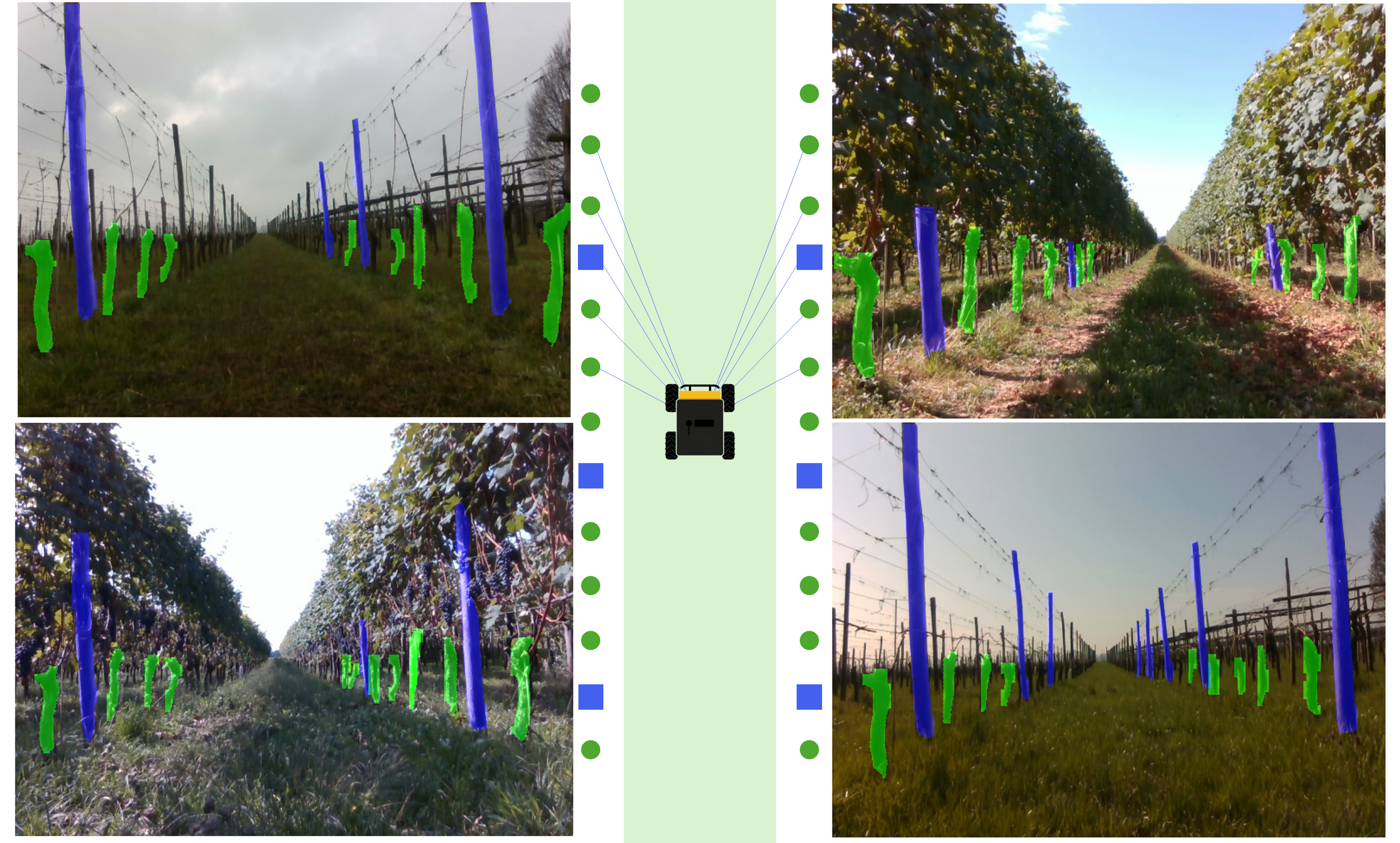}
    \caption{VinePT-Map enables robotic platforms to build permanent landmarks map in vineyards for robust autonomous operation in all the seasons.}
    \label{fig:first-page}
\end{figure}

Vineyards exemplify these complexities. Their highly repetitive row geometry gives rise to perceptual aliasing \cite{3DMoveToSee}, while fluctuations in canopy density, illumination conditions, and phenological stages significantly transform the environment’s visual characteristics over time \cite{devInAgri}. Therefore, mapping and localization strategies that depend on generic visual features are inevitably susceptible to robustness issues, which restricts their suitability for sustained, long-term deployment. As a result, the applicability of canopy-centric localization and mapping methodologies is limited to short-term data acquisition campaigns and cannot generalize to extended time horizons and to diverse environmental conditions \cite{RevCanopy} \cite{ChangingOrchards}. 
Recent advances in mobile robotics have underscored the advantages of incorporating structural or semantic landmarks for navigation \cite{desilva2025} \cite{agriPlane}. 
This work focuses on the idea of a semantic mapping framework for vineyards based on persistent landmarks, specifically vine trunks and support poles. These structural elements of the field provide a robust and season-agnostic spatial foundation for accurate localization and mapping of autonomous robots.
The proposed approach mirrors fundamental principles of human spatial cognition: humans navigate complex environments by forming cognitive maps anchored to stable, semantically meaningful landmarks, rather than relying on superficial visual appearances that may vary \cite{Epstein2017}. Similarly, rather than relying on the variable visual cues of foliage or grass, which undergo seasonal changes, this work relies on the vineyard's permanent skeletal infrastructure.

We propose VinePT-Map: a poles-trunks semantic mapping framework based on factor graphs. This graphical representation allows for the seamless fusion of GPS-RTK and RGB-D data, posing geometrical constraints to exploit the structure of the field. By decoupling variables and measurements, the approach naturally supports incremental refinement of object poses and map consistency, while maintaining robustness to sensor noise and partial observations. The perception entails an instance segmentation and tracking stage, a precise reference point computation for map projection, and a landmark data association step for robust outlier rejection. The overall system can be deployed directly on the robot with low-power consumption and low-cost hardware, building the map of the field at runtime. An extensive validation has been performed to test the perception and mapping algorithms across all the seasons in a real vineyard, as seen in Fig.~\ref{fig:first-page}. We validate the visual perception with a new custom dataset for instance segmentation and tracking of poles and trunks, while the mapping results have been quantitatively computed using georeferenced positions of the landmarks. The implementation and the access to data will be publicly released in the VinePT-Map repository \footnote{https://github.com/PIC4SeR/VinePT-Map}.

The contributions of the paper can be summarized as:
\begin{itemize}
    \item VinePT-Map: a poles-trunks semantic mapping methodology based on factor graphs to map permanent structural elements, reducing map complexity and increasing resilience to environmental aliasing using consumer-grade RGB-D cameras;
    \item A dataset to train and test instance segmentation and tracking of poles and trunks, spanning from February to September;
    \item An efficient segmentation and tracking pipeline for poles and trunks, coupled with a clustering filter to reject outliers, that can incrementally create a map while the robot navigates through the field over the entire year;
    \item A thorough experimental evaluation on the field of both perception and mapping performance across seasons and varying vineyard conditions, and an ablation study highlighting the impact of perceptual methods on mapping performance.
    
\end{itemize}

\section{Related Works}
\label{sec:related_works}
Autonomous navigation in precision viticulture faces critical challenges from perceptual aliasing and seasonal morphological volatility, thus making traditional geometric mapping inadequate \cite{3DMoveToSee}\cite{desilva2025}. State-of-the-art research is shifting towards the integration of semantic-aware Simultaneous Localization and Mapping (SLAM) systems, with the objective of leveraging semantic landmarks to effectively address the geometric ambiguities inherent in repetitive crop rows \cite{desilva2025} \cite{rapado2025tree}.
This paradigm relies heavily on edge-deployed deep learning architectures, such as YOLO \cite{yolo}, to achieve real-time feature extraction in unstructured environments. In order to exceed the constraints imposed by purely visual systems, these semantic detections are combined with 3D LiDAR data \cite{Moreno2020-bf}.
Recently proposed SLAM frameworks capitalize on these detections to model agricultural environments robustly \cite{aguiar2020localization}. For instance, VineSLAM \cite{agriPlane} employs semi-plane segmentation to map parallel rows, while Tree-SLAM \cite{rapado2025tree} treats individual trunks as discrete entities within factor graphs. Furthermore, the introduction of "semantic walls" has been shown to act as a pseudo-rigid constraint within semantic particle filters, leading to a significant reduction in longitudinal drift along symmetrical corridors \cite{agriPlane}.
In order to ensure temporal invariance and enable year-round operation despite drastic canopy variations, recent approaches employ topological neighborhood descriptors and multi-metric tracking frameworks for robust data association \cite{rapado2025tree} \cite{JIANG2024108870}. In view of the current state of the art, we posit that leveraging permanent semantic features, such as poles and trunks, rather than raw geometric data, provides the requisite precision for complex, long-term autonomous operation.

\section{Methodology}
\label{sec:methodology}

\begin{figure*}[ht]
    \centering
    \includegraphics[width=\textwidth, angle = 0]{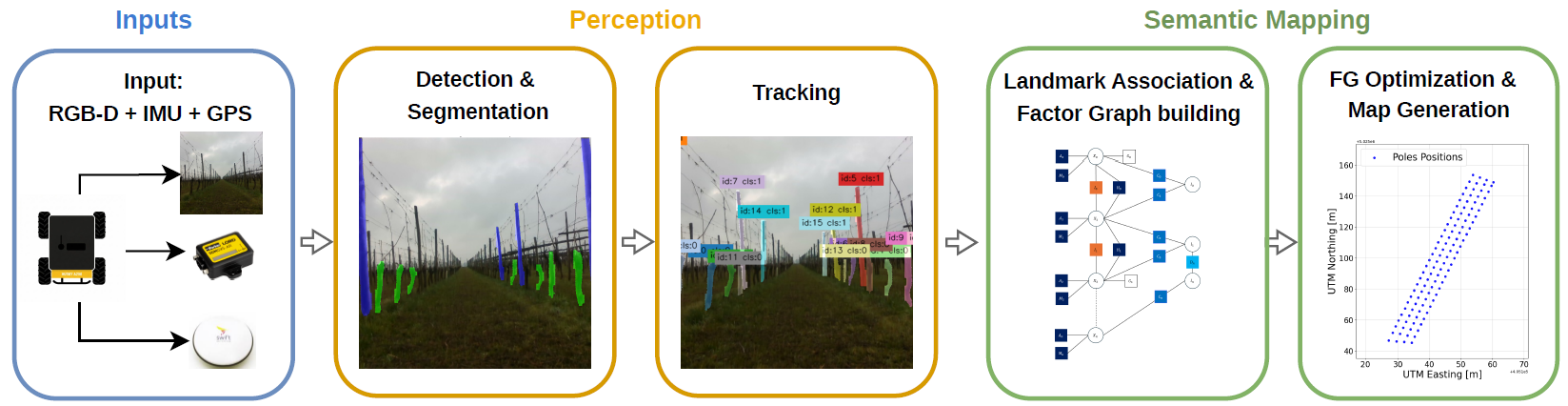}
    \caption{VinePT-Map: a schematic pipeline of the semantic mapping framework for persistent landmarks in vineyards.}
    \label{fig:schema}
\end{figure*}

This section presents the pipeline for building a georeferenced, semantically annotated map of the structural elements of vineyards. The proposed pipeline transforms raw RGB-D images, IMU and GPS measurements into a globally consistent set of landmark positions, each annotated with a semantic class label and an associated uncertainty estimate.

\subsection{Methodology overview}
 The system is organized into a two cascaded stages reported in Fig.\ref{fig:schema}. The first stage, the \textit{perception front-end}, operates on each RGB-D frame pair to segment, and track vineyard  vine trunks and support poles in order to estimate their positions in camera reference frame. The second stage, the \textit{mapping back-end}, fuses the resulting landmark observations with IMU and GPS measurements into a joint maximum a posterior (MAP) estimation problem over the robot trajectory and all landmark positions, solved incrementally via factor graph optimization. A deferred data-association strategy bridges the two stages, ensuring that only filtered, outlier-free observations are committed to the graph. 

Let $\mathcal{X} = \{\mathbf{T}_k, \mathbf{v}_k\}_{k=0}^{K}$ denote the robot trajectory states and $\mathcal{L} = \{\boldsymbol{\ell}_j\}_{j=1}^{J}$ the landmark map. The complete estimation problem seeks the joint MAP estimate
\begin{equation}
\mathcal{X}^*,\, \mathcal{L}^* = \arg\max_{\mathcal{X},\, \mathcal{L}} \; p(\mathcal{X}, \mathcal{L} \mid \mathcal{Z})
\end{equation}
where $\mathcal{Z}$ aggregates all sensor measurements. Under the standard assumption of Gaussian noise and conditional independence given the variables each factor connects.

\subsection{Segmentation and Tracking Dataset}
Trunks and poles segmentation requires sufficient data to train neural network models. Unfortunately, those classes are not commonly included in public computer vision datasets. Hence, a significant data collection on the field has been carried out to perform the study. A new dataset has been gathered using an Intel Realsense D435 camera with resolution $640\times480$ pixels, mounted frontally on a ClearPath Husky rover.
To promote multi-season generalization of the segmentation model, multiple data collection campaigns have been performed over the course of the entire year, capturing the vineyard across distinct phenological stages, from February to September, to account for significant variations in foliage density, grass growth, and illumination.
The images have been manually labeled with ground truth masks, totally comprising 1600 RGB-D images. Moreover, a ground truth has been generated for tracking the instances of poles and trunks. For each frame of the video stream, a coherent ID has been assigned to each instance to precisely validate the performance of the perception pipeline along the entire experiment duration.

\subsection{Perception Front-End}
\label{sssec:Front-End}
The perception front-end receives time-synchronized colour images $\mathbf{I}_t \in \mathbb{R}^{H \times W \times 3}$ and aligned depth maps $\mathbf{D}_t \in \mathbb{R}^{H \times W}$ acquired by an RGB-D camera at each discrete time step $t$. Its output is a set of per-frame landmark descriptors:
$$
\mathcal{O}_t = \left\{ \left( \mathbf{p}_i^{\,c},\; s_i,\; c_i,\; \tau_i \right) \right\}_{i=1}^{N_t}
$$

where $\mathbf{p}_i^{\,c} \in \mathbb{R}^3$ is the estimated centroid in the camera frame, $s_i \in [0,1]$ is the detection confidence, $c_i \in \{0, 1\}$ is the semantic class label (trunk or pole), and $\tau_i \in \mathbb{Z}$ is the persistent tracking identity assigned by the multi-object tracker. The processing chain is detailed in the following subsections.\\

\noindent\textbf{Instance Segmentation and Tracking:} The perception front-end employs a \textit{YOLOv8-seg} architecture \cite{yolo} for real-time segmentation. The model was fine-tuned on the custom multi-temporal vineyard dataset collected. 

The dataset is partitioned into a training set ($75\%$), a validation set ($10\%$) and a testing set ($15\%$) to ensure an unbiased performance evaluation. For each input frame $\mathbf{I}_t$, the model extracts two classes: trunks ($c = 0$) and poles ($c = 1$). Each detection $\mathbf{d}_i$ is defined by a binary mask $\mathbf{M}_i$, a class prediction $c_i$, and a confidence score $s_i$. To ensure high precision, we discard detections where:

\begin{equation}
    s_i < \theta_{\mathrm{conf}} \quad \text{or} \quad c_i \notin \mathcal{C}_{\mathrm{allowed}}
\end{equation}

where $\theta_{\mathrm{conf}}=0.8$ is the minimum confidence threshold used and $\mathcal{C}_{\mathrm{allowed}}$ is the set of permitted class identifiers. Finally, Non-Maximum Suppression is applied using an Intersection-over-Union (IoU) threshold $\theta_{\mathrm{IoU}}$ to eliminate redundant predictions.

In order to ensure landmark ID temporal consistency across the image sequence, the segmentation output is integrated with the BoT-SORT \cite{botsort} multi-object tracking framework. The purpose of this module is to enable the transition between individual frame detections and persistent landmark identification in image frames. To each valid detection is assigned a unique, persistent integer track identity, denoted by the symbol $\tau_i$. This identifier is the cornerstone of the system's data-association mechanism. By treating all observations with the same value of the variable $\tau_i$ as a single physical entity, the system eliminates the computational overhead of visual re-identification in subsequent mapping and optimization stages.

\noindent\textbf{Landmark Projection in 2D map:} For each retained detection $i$ with binary mask $\mathbf{M}_i$, the corresponding region in the aligned depth map $\mathbf{D}_t$ is extracted, and the masked depth pixels are back-projected into a three-dimensional point cloud in the camera coordinate frame $\{C\}$. The back-projection follows the pinhole camera model. A validity filter rejects pixels whose depth values fall outside the admissible range $[z_{\min}, z_{\max}]$ or are non-finite. Masks yielding fewer than $N_{\min}$ valid three-dimensional points are discarded to suppress detections originating from distant or poorly resolved objects. The resulting per-instance point cloud is denoted $\mathcal{P}_i = \{ \mathbf{p}_n^c \}_{n=1}^{M_i} \subset \mathbb{R}^3$.

\noindent\textbf{Landmark reference point computation:} The reference position of each detected object is estimated from its reconstructed point cloud $\mathcal{P}_i$. The landmark spatial coordinates are derived through a localized centroid estimation that focuses exclusively on the lower segment of the detected object. By prioritizing this structurally rigid region, the system mitigates the impact of transient environmental noise, thereby facilitating more reliable data association and drift reduction within the subsequent mapping framework.

The isolation of the structurally stable base is implemented by sorting the landmark point cloud $\mathcal{P}_i$ along the vertical camera axis ($y$-axis, oriented downward). To define the base subset $\mathcal{P}_i^b$, a trimming operation is applied whereby only the lowest fraction, specifically the first quarter, of the vertical distribution is retained.  
The 3D coordinates of the landmark reference point are subsequently derived via a hybrid mean--median estimator, formulated to balance precision with robustness:
\begin{equation}
    \bar{\mathbf{p}}_i^{\,c} = \left[ \operatorname{mean}(\mathcal{P}_{i,x}^b), \; \operatorname{mean}(\mathcal{P}_{i,y}^b), \; \operatorname{median}(\mathcal{P}_{i,z}^b) \right]^\top
\end{equation}
The lateral coordinates ($x, y$) are derived via the arithmetic mean, a choice that is substantiated by the relatively low noise levels in the lateral image plane. In contrast, the depth coordinate ($z$) is determined using a median estimator. This provides critical resilience against heavy-tailed noise distributions and depth discontinuities that characterize depth measurement from RBG-D cameras in outdoor scenarios. The decoupling of the coordinate estimation in this way enables the system to achieve the statistical robustness required for unstructured environments while maintaining a lightweight computation profile.

\begin{figure}[ht]
    \centering
    \includegraphics[width=0.9\columnwidth, angle = 0]{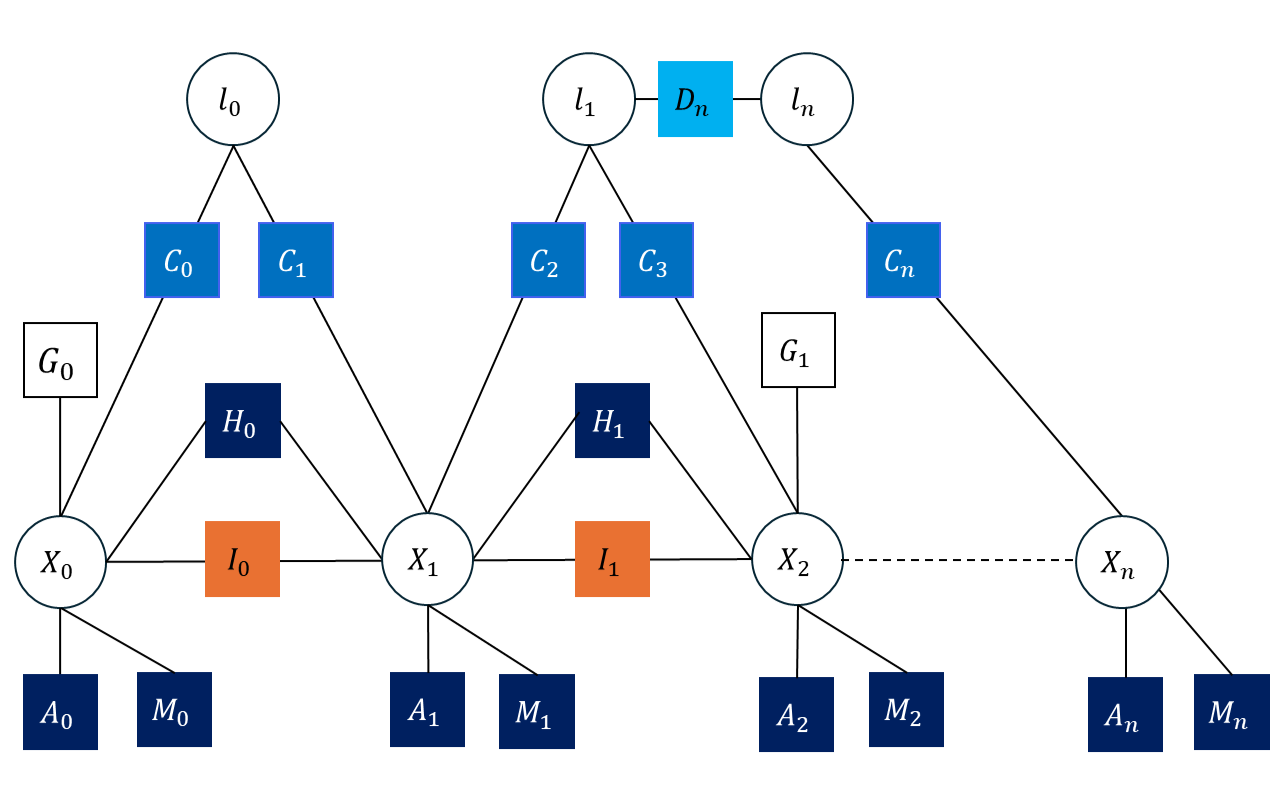}
    \caption{VinePT-Map Factor Graph for poles and trunks mapping.}
    \label{fig:factor_graph}
\end{figure}

\subsection{Localization and Mapping with Factor Graph}
The back-end fuses landmark observations from the perception front-end \ref{sssec:Front-End} with inertial and GNSS measurements to jointly estimate the robot trajectory and a georeferenced landmark map. The estimation is cast as MAP inference on a factor graph $\mathcal{G} = (\mathcal{V}, \mathcal{F})$, where $\mathcal{V}$ denotes variable nodes and $\mathcal{F}$ the set of factor (constraint) nodes, and is solved incrementally with the iSAM2 algorithmn\cite{isam2}. Fig. \ref{fig:factor_graph} illustrates the factor graph structure. Section \ref{subsub:1} presents the robot localisation formulation; Section \ref{subsub:2} describes the data-association strategy; Section \ref{subsub:3} details landmark mapping and post-hoc map refinement.

\subsubsection{Robot Localization}
\label{subsub:1}
At each keyframe $k = 0, \dots, K$ the robot state comprises its six-degree-of-freedom pose $\mathbf{T}_k \in SE(3)$, linear velocity $\mathbf{v}_k \in \mathbb{R}^3$. Graph construction is deferred until a valid GNSS fix and a calibrated AHRS orientation are simultaneously available.
A set of factors constrains the robot's state at each keyframe:

\noindent\textbf{IMU Preintegration:} Between consecutive keyframes, high-rate accelerometer and gyroscope readings are compressed into a single on-manifold pre-integrated measurement factor $I_n$ following \cite{IMU_pre}. 

\noindent\textbf{RTK-GPS Position:}
Each valid RTK-GPS fix constrains the translational component of the pose after conversion to ENU coordinates $G_n$.

\noindent\textbf{Attitude :} A two-dimensional factor $A_n$  constrains roll and pitch by comparing the body-frame gravity projections of the estimated and AHRS-measured orientations, while leaving yaw unconstrained.

\noindent\textbf{Magnetometer Heading:} Absolute heading is maintained via a wrap-free parameterization. By formulating the residual of factor $M_n$ using the rotation matrix columns, the system avoids discontinuities at $\pm\pi$, ensuring smooth optimization gradients.

\noindent\textbf{Non-Holonomic Constraints:} To exploit the known kinematics of the rover, a soft factor $H_n$ penalizes lateral and vertical velocities. A Huber robust cost is employed to maintain estimation integrity during maneuvers on slippery vineyard soil.

At each keyframe, the accumulated factors are committed to the iSAM2 optimiser \cite{isam2}, which performs partial re-elimination of its Bayes-tree representation, achieving amortised constant-time updates. 

\subsubsection{Landmark Data Association}
\label{subsub:2}
This step of the pipeline addresses landmark association. The tracker identity $\tau_i$ assigned during the perception stage provides the primary short-term correspondence. The proposed system addresses global landmarks association through two complementary stages: deferred commitment with statistical outlier rejection, and class-aware spatial merging.

\noindent\textbf{Deferred Commitment:} Rather than being inserted into the factor graph immediately landmarks observations are buffered per tracker identity. Each buffered record contains the associated robot pose key, the bearing–range measurement with its propagated noise model, the detection confidence, and an approximate world-frame position $\tilde{\boldsymbol{\ell}} = \mathbf{T}_k \circ \mathbf{p}^b$. The landmark commitment is triggered when the landmark is determined to be outside the current sensor field of view.

\noindent\textbf{Outlier Rejection:}Prior to commitment to optimization, the buffered world-frame landmark estimates $\{\tilde{\boldsymbol{\ell}}^{(n)}\}$ are filtered using the Median Absolute Deviation (MAD). Observations whose normalized deviation $z_n = \lVert \tilde{\boldsymbol{\ell}}^{(n)} - \tilde{\boldsymbol{\ell}}_{\mathrm{med}} \rVert / (1.4826 \cdot \mathrm{MAD})$ exceeds the threshold $\lambda_{\mathrm{MAD}}$ = 1.5 are discarded.

\noindent\textbf{Class-Aware Spatial Merging:} The initial landmark position is set to the component-wise median of the inlier set: $\boldsymbol{\ell}_j^{(0)} = \mathrm{median}(\{\tilde{\boldsymbol{\ell}}^{(n)}\}_{n \in \mathrm{inliers}})$. Before a new variable is inserted, a nearest-neighbour search is performed among existing landmarks of the same semantic class. If an existing landmark $\boldsymbol{\ell}_{j'}$ satisfies $\lVert \boldsymbol{\ell}_j^{(0)}{}_{xy} - \boldsymbol{\ell}_{j'}{}_{xy} \rVert < d_{\mathrm{merge}}$ the incoming observations are redirected to $\boldsymbol{\ell}_{j'}$, resolving fragmented tracks caused by occlusion-induced breaks without introducing duplicate variables. Restricting the search to same-class landmarks prevents spurious merges across semantically distinct object categories. $d\_merge$ holds two discrete values for each category: $0.5 m$ for trunks and $1 m$ for poles.

\subsubsection{Landmark Mapping}
\label{subsub:3}
In conclusion, the estimation of pole and trunk positions is achieved by the integration of landmark observations within the optimisation problem.

\noindent\textbf{Bearing–Range Observation Model:} Each committed observation, originally a Cartesian point $\mathbf{p}^b$ in the body frame, is converted to a bearing $\mathbf{u} = \mathbf{p}^b / \lVert \mathbf{p}^b \rVert \in \mathbb{S}^2$ and range $r = \lVert \mathbf{p}^b \rVert$. The measurement covariance in bearing–range space is obtained by propagating the Cartesian sensor noise $\boldsymbol{\Sigma}_{xyz}$ through the transformation Jacobian:
$$
\boldsymbol{\Sigma}_{\mathrm{BR}} = \mathbf{J} \, \boldsymbol{\Sigma}_{xyz} \, \mathbf{J}^\top, \qquad \mathbf{J} = \begin{bmatrix} \mathbf{B}^\top \, \frac{1}{r}(\mathbf{I}_3 - \mathbf{u}\mathbf{u}^\top) \\[2pt] \mathbf{u}^\top \end{bmatrix}
$$
where $\hat{\mathbf{u}}$ and $\hat{r}$ are the predicted bearing and range, and $\mathbf{B}(\hat{\mathbf{u}}) \in \mathbb{R}^{3 \times 2}$ is an orthonormal basis spanning the tangent plane at $\hat{\mathbf{u}}$ on $\mathbb{S}^2$.
The resulting covariance is symmetrized and regularized to guarantee positive definiteness.

\noindent\textbf{Map Refinement:} Upon completion of the traverse, a consolidation step addresses residual landmark map fragmentation. Landmark positions are clustered per semantic class using DBSCAN with neighbourhood radius $\epsilon_{\mathrm{cluster}}$.The $\epsilon_{\mathrm{cluster}}$ is computed as one-half of the mean distance between objects of the same class that have been mapped. For each pair $(\boldsymbol{\ell}_a, \boldsymbol{\ell}_b)$ within a cluster, a zero-displacement between-factor $D_n$ is added.
These soft constraints attract co-located landmarks toward a consensus position without irreversibly discarding variables. The complete graph is then re-optimized, yielding the final landmark map $\{\boldsymbol{\ell}_j^*\}$ and robot trajectory $\{\mathbf{T}_k^*\}$ in the ENU frame, each landmark annotated with its semantic class $c_j \in \{\text{trunk},\, \text{pole}\}$.

\input{campaign_table}

\section{Experiments}
Four field campaigns have been carried out to collect the data and trajectories of the robot in different seasons. Table \ref{tab:campaign} reports the field and weather conditions of each campaign, affecting the performance of perception and SLAM systems.
The experiments has been performed with a Husky A200 rover from ClearPath Robotics (size 990 x 670 x 390 mm) mounting an on-board computer with an Intel i7-6700E processor, 16 GB DDR4 RAM. The adopted sensor suite includes an RGB-D Intel Realsense D435 camera, providing aligned RGB and depth images, an accurate AHRS MicroStrain 3DM-GX5, and a RTK-GPS.

\begin{figure}[t]
    \centering
    \includegraphics[width=\columnwidth]{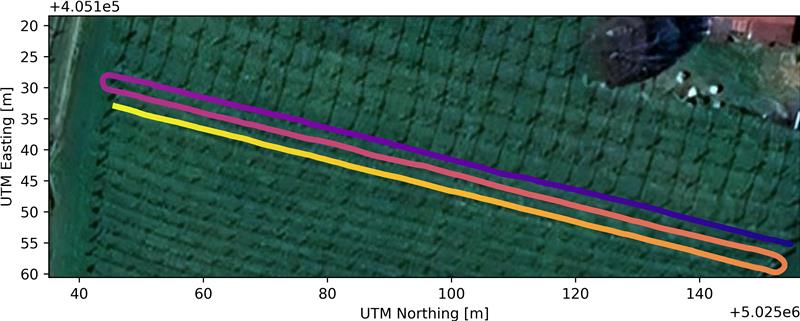}
    \caption{Aerial view of the vineyard with the trajectory of the robot over three rows (starting from dark line, ending in yellow line), used to compute the incremental result of Table~\ref{tab:table_map2}.}
    \label{fig:satellite_view}
\end{figure}

The rover has been teleoperated with  $v_{max}=1~m/s$ and $ \omega_{max}=2.0~rad/s$ following the desired trajectory along three consecutive rows of the vineyard, about $100~m$ long, shown in Fig \ref{fig:satellite_view}. In some trajectory segments, the difficulty of remote driving, the terrain irregularity, and the small obstacles led to sudden drifts of the rover, missing some landmarks with the camera on one side. This behavior can also occur in realistic autonomous driving settings.
The ground truth position of the pole landmarks on the field has been obtained with an aerial georeferenced image.


\begin{table}[t]
\centering
\caption{Results of landmarks instance segmentation and tracking. The best seasonal results are highlighted in bold.}
\label{tab:table-seg}
\resizebox{\columnwidth}{!}{%
\begin{tabular}{@{}ccccll@{}}
\toprule
Class       & Date    & Box mAP50      & Mask mAP50     & HOTA          & AssA          \\ \midrule
       & 02-12   & 0.934          & 0.928          & 89.1          & \textbf{90.3} \\
       & 03-20   & \textbf{0.961} & \textbf{0.946} & \textbf{85.4} & 89.8          \\
\textbf{Trunks} & 08-14   & 0.91           & 0.916          & 74.7          & 87.1          \\
       & 09-11   & 0.863          & 0.859          & 81.3          & 84.8          \\ \cmidrule(l){2-6} 
       & Overall & 0.914          & 0.908          & 82.6          & 88.0          \\ \midrule
       & 02-12   & 0.878          & 0.833          & \textbf{72.4} & 74.7          \\
       & 03-20   & 0.895          & 0.823          & 70.8          & 72.5          \\
\textbf{Poles}  & 08-14   & \textbf{0.913} & \textbf{0.902} & 51.7          & \textbf{95.1} \\
       & 09-11   & 0.828          & 0.855          & 67.6          & 74.9          \\ \cmidrule(l){2-6} 
       & Overall & 0.887          & 0.844          & 65.6          & 79.3          \\ \bottomrule
\end{tabular}%
}
\end{table}

\begin{table}[t]
\centering
\caption{Quantitative evaluation of poles landmarks mapping using georeferenced positions as ground truth. The metrics are reported incrementally over three consecutive rows.}
\label{tab:table_map2}
\resizebox{\columnwidth}{!}{%
\begin{tabular}{@{}lcc|cc|cc@{}}
\toprule
N. Rows & \multicolumn{2}{c|}{1} & \multicolumn{2}{c|}{2} & \multicolumn{2}{c}{3} \\ \midrule
Date    & MAE [m]$\downarrow$    & TP$(\%)\uparrow$     & MAE [m]$\downarrow$      & TP$(\%)\uparrow$     & MAE [m]$\downarrow$      & TP$(\%)\uparrow$  \\ \midrule
02-11   & 0.20       & \textbf{100.0}          & \textbf{0.18}       & 97.3          & \textbf{0.18}       & 97.1         \\
03-20   & 0.24       & 96.2          & 0.22       & 96.3          & 0.27       & 91.1         \\
08-14   & 0.25       & 98.1          & 0.24       & 97.2         & 0.24      & 93.1         \\
09-11   & 0.31      & \textbf{100}          & 0.29       & 96.3          & 0.26       & 97.2         \\ \bottomrule
\end{tabular}%
}
\end{table}

\section{Results}
\label{sec:results}
In this section, the resulting effectiveness of the VinePT-Map system has been evaluated on both perception and mapping aspects. First, we measured the reliability of the detection and tracking pipeline on our custom image dataset spanning all the seasons, to evaluate how well it identifies and tracks poles and trunks. Hence, the generalization of the detection model over the different conditions has been analyzed. Then, the spatial accuracy of the semantic maps obtained through the factor graph optimization is discussed under different phenological vegetation stages.

\subsection{Visual Perception Evaluation}
A major challenge for agricultural robots is maintaining a reliable perception of relevant features, even when the environment changes drastically throughout the year. We evaluated our perception pipeline using three metrics: Mean Average Precision (mAP50), Higher Order Tracking Accuracy (HOTA) and Association Accuracy (AssA). As shown in Table \ref{tab:table-seg}, the system performed consistently across all temporal evaluations.

\textbf{Detection:} During the winter and early spring (February and March), when the vines were bare stems, trunk detection was most effective, achieving Box mAP50 scores of 0.934 and 0.961, respectively. Importantly, as the canopy transitioned to the full foliage and fruiting stages in August and September, the framework maintained high detection rates, achieving Box mAP50 scores of 0.910 and 0.863, respectively. This marginal variance empirically confirms that detection can perform consistently throughout each stage of phenological growth. Detection and segmentation performances have not shown performance degradation even in more complex seasons such as summer and autumn, when foliage introduces more noise and covers part of the poles and trunks. In addition, our solution has proven robustness also to different lighting conditions due to the weather and light direction.

\textbf{Tracking:} Persistent tracking is paramount to mitigate the perceptual aliasing inherent in repetitive crop rows. The system yielded an overall AssA of 88.0 for trunks and 79.3 for poles. Notably, the pole tracking module achieved an exceptional AssA peak of 95.1\% during the complex summer traverse in August, demonstrating the tracker’s ability to maintain consistent tracking IDs despite the dense vegetation and changing lighting conditions. The overall results of tracking can be partly attributed to the great performance of the detection algorithm. In fact, stable detections over time pave the way for a more stable functioning of the tracking algorithm. Moreover, due to the limited movements of the tracked object, the tracker is able to provide more robust results.

\textbf{Inference:} The real-time capability of the perception pipeline is fundamental to be deployed directly on the autonomous platform. Hence, the segmentation and tracking pipeline has been tested for a real test on the robot computational hardware. On the i7-6700E processor of the robot board, using an image resolution of $640 \times 480$ px, the segmentation model and tracking pipeline can run at 30 Hz, which is the acquisition frequency of the camera.

\subsection{Mapping Evaluation}
A quantitative evaluation of the performance of the VinePT-Map semantic framework is proposed here. The geometrical accuracy of the generated poles landmarks maps is directly computed. We compare the computed maps, optimized by the factor-graph methodology, against georeferenced ground truth data acquired through precise aerial images and photogrammetry software. The adopted metrics are the Mean Absolute Error (MAE) to directly compare the estimate with the ground truth value, and the percentage of landmarks mapped correctly. The results are reported in Table \ref{tab:table_map2}.

The mapping pipeline demonstrated high precision, particularly during the winter campaign in February, when optimal visibility is achieved without foliage. It recorded a MAE of 0.18 m when mapping two or three consecutive rows. In August, with the most challenging conditions of both grass and canopies, the maximum error marginally increased to only 0.32 m. This result confirms our hypothesis that poles and skeletal elements of the field remain visible; thus, they represent persistent landmarks. Moreover, the depth-sensor noise induced by thick vegetation's occlusions can be effectively bounded by the factor graph's robust geometrical constraints. 
The percentage of True Positive (TP)$\%$ in Table \ref{tab:table_map2} evaluates the number of poles landmarks correctly mapped and re-associated over multiple observations.
The system's ability to persistently capture the topology of the vineyard is reflected in its high landmark mapping rates. During the winter and summer surveys, the percentage of successfully mapped poles consistently exceeded 93.1\%, reaching 100\% in the primary row in February. 

\begin{figure}[t]
   \centering
   \includegraphics[width=\columnwidth, angle = 0]{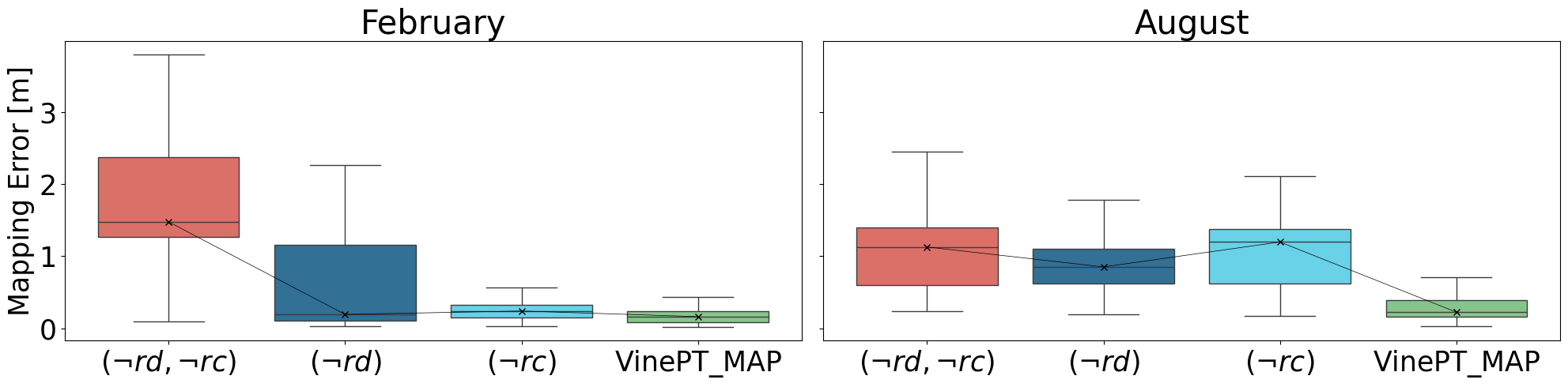}
   \caption{Box plots of the ablation study results on poles mapping error. The relevance of the landmark reference point computation method (rd) and the data association method (rc) is highlighted by excluding them from the pipeline.}
   \label{fig: Ablation}
\end{figure}

\begin{figure*}[t]
    \centering
    \setlength{\tabcolsep}{0pt}
    \centering
    \begin{tabular}{c c}
    \includegraphics[width=0.34\linewidth]{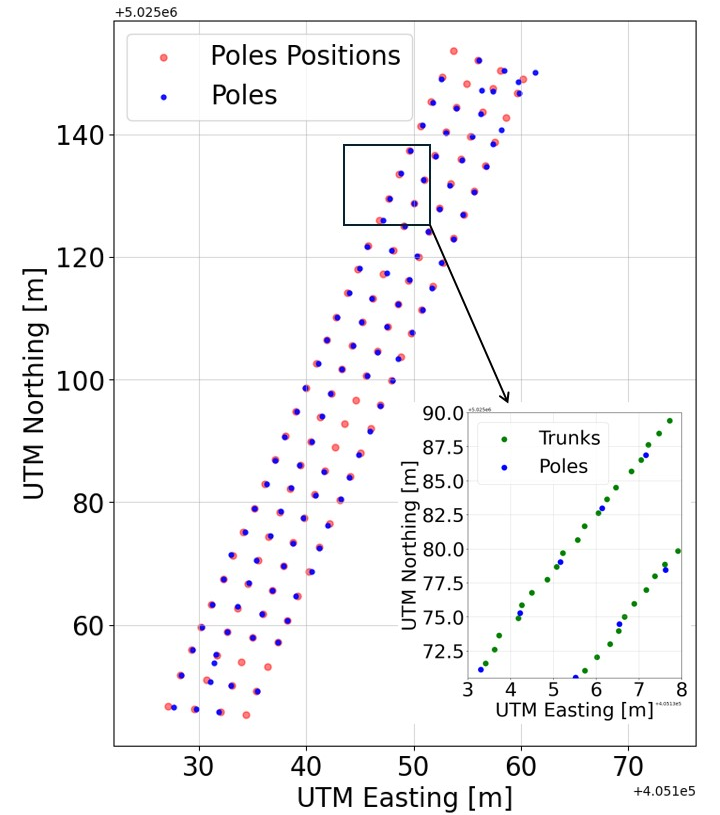} & \includegraphics[width=0.344\linewidth]{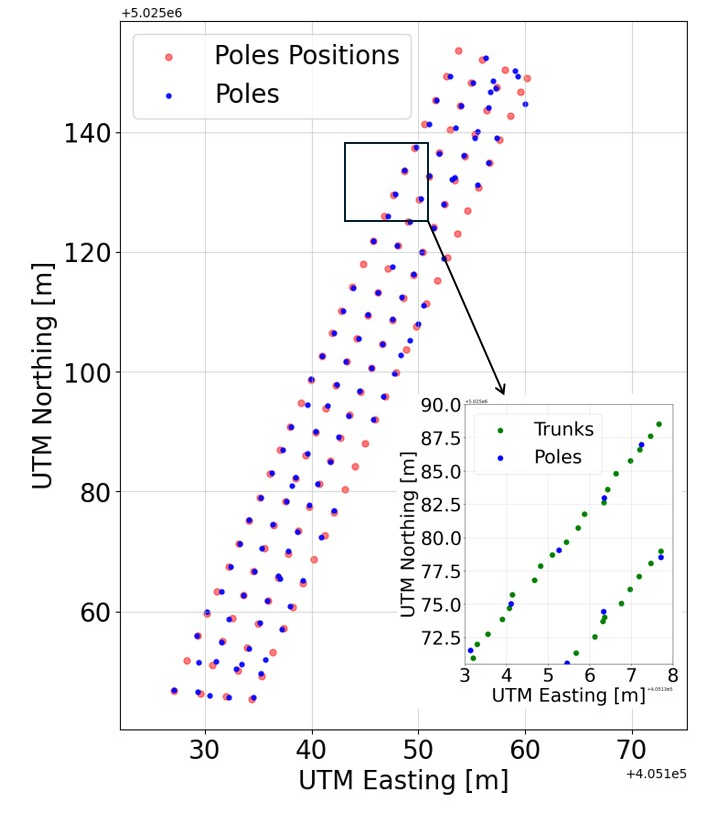} \\
    (a) February & (b) March\\
    \includegraphics[width=0.34\linewidth]{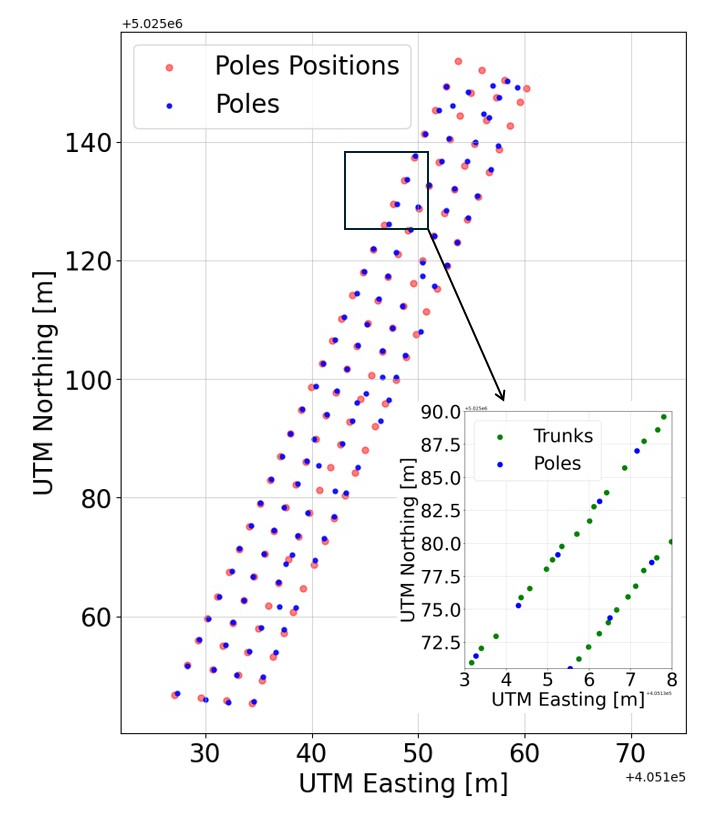} & \includegraphics[width=0.319\linewidth]{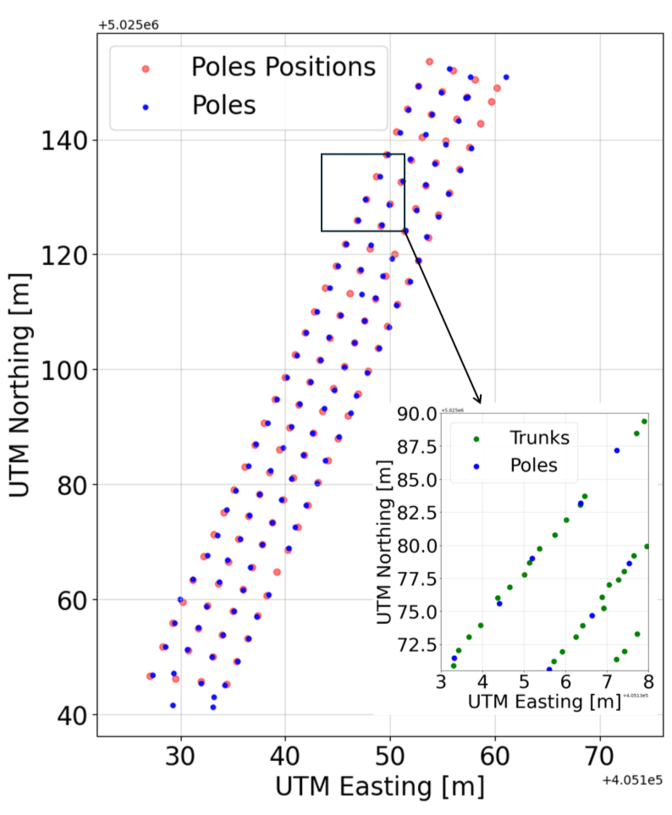} \\
    (c) August & (d) September\\
    \end{tabular}
    \caption{Map of vineyard poles and trunks over three consecutive rows in February, March, August, and September, with a zoomed-in view of the poles and trunks.}
    \label{fig:maps_result}
\end{figure*}

\subsection{Ablation study}
An ablation study is performed to investigate the impact of fundamental perception steps on the mapping results. The box plots of poles mapping error over three consecutive vineyard rows, for the field campaign of February and August, are reported in Fig.\ref{fig: Ablation}. The performance of the VinePT-Map system is analyzed, removing the landmark reference point computation method (rd) and the deferred commitment with MAD-based outlier rejection (rc), explained in Section \ref{sec:methodology}.
In February, with leafless canopies and structural elements fully exposed, the results clearly demonstrate the positive impact of each component alone. In this scenario, both contributions independently yield substantial improvements in mapping accuracy, reducing the median error from $1.4\,\text{m}$ in the baseline configuration $(\neg rc,\, \neg rd)$ to $0.16\,\text{m}$ in the full VinePT-Map pipeline. In these field conditions, the robust reference-point extraction method ($rd$) significantly contributes to achieving accuracy comparable to the complete system. However, under the more challenging conditions of August, with thick canopies and fruits obstructing the measurement process, the same trend does not hold. Under these adverse conditions, neither component acting independently is sufficient to produce a meaningful improvement over the baseline: the median error and variance of the mapping error remain elevated across both partial configurations $(\neg rc)$ and $(\neg rd)$. Only the full VinePT-Map pipeline, integrating both contributions, achieves a median error of $0.22\,\text{m}$. These ablation results provide empirical evidence that the two proposed components are jointly necessary to ensure robust and consistent landmark mapping performance across the full range of seasonal conditions encountered in real-world vineyard environments.

\subsection{Qualitative Map Evaluation}
The semantic maps of poles, landmarks, and associated ground truth are shown in Fig. \ref{fig:maps_result}. The zoom-in view also shows the trunk's position estimated for a certain portion of the trajectory. In general, the highest uncertainty in landmarks mapping is localized at the end of the rows, where the rover performs a strict turn, losing some poles, and plants may present irregular dispositions. In February, a series of poles in the central part of the map, second row, were removed for substitution and have not been mapped by the robot. In March, the map was correctly updated. On the other hand, some poles have been missed on the third row (March) due to abrupt heading changes and oscillations of the robot. August presents the most frequent re-association and precision errors, being the most challenging scenarios with thick canopies and grapes.

\section{Conclusions}
\label{sec:conclusions}
This work presented VinePT-Map, a semantic mapping framework for vineyards that leverages persistent structural landmarks to enable robust and season-agnostic localization of autonomous robots. By formulating the problem within a factor graph framework and fusing robot localization and RGB-D measurements through structural geometric constraints, the proposed approach reduces susceptibility to perceptual aliasing and environmental variability.

The integration of an efficient instance segmentation and tracking pipeline, coupled with a clustering-based refinement strategy, allows reliable landmark detection using low-cost sensors and onboard computation, supporting practical field deployment. Extensive multi-season experiments, together with the introduction of a dedicated dataset for poles and trunks, validate the robustness of both perception and mapping components under varying illumination, canopy density, and weather conditions.

Overall, the results demonstrate that semantic vineyard maps based on permanent skeletal infrastructure can be obtained with high precision in different seasons and field conditions, providing a stable spatial foundation for long-term autonomous operation. The ablation study remarked the importance of a robust perception pipeline to obtain reliable field maps. Future work will be focused on the extension of the experimentation to vineyards of different types, collecting also the georeferenced ground truth of trunk landmarks. A pivotal step will be the integration of the semantic map obtained into a landmark-based localization system, comparing the performance with GPS localization and exploring the reliability of low-cost visual solutions. Moreover, the semantic skeletal map, target of this study, constitutes a first ground layer of a broader semantic field description to be build on top, comprising fruits poses and features that will be the main focus of agricultural robots activity.


\bibliographystyle{IEEEtran}
\bibliography{bibliography}

\end{document}

%% file: campaign_table.tex
\begin{table}[ht]
\centering
\caption{Overview of field campaigns in the vineyards.}
\label{tab:campaign}
\resizebox{\columnwidth}{!}{%
\begin{tabular}{@{}lccl@{}}
\toprule
\textbf{Campaign} & \textbf{Weather} & \textbf{Plant Growth} & \textbf{Grass Height} \\ \midrule
12/02 &   Cloudy       & Stem, no leaves             & Low ($\sim 5$ cm)    \\
20/03 &   Sunny          & Stem, no leaves             & Medium ($\sim 20$ cm)    \\
14/08 &   Sunny         & Branches, leaves, fruit     & Low ($\sim 5$ cm)    \\
11/09 &   Cloudy       & Branches, leaves, fruit     & Medium ($\sim 20$ cm) \\
\bottomrule
\end{tabular}%
}
\end{table}

%% file: bibliography.bib
@article{rapado2025tree,
  title={Tree-SLAM: semantic object SLAM for efficient mapping of individual trees in orchards},
  author={Rapado-Rincon, David and Kootstra, Gert},
  journal={Smart Agricultural Technology},
  pages={101439},
  year={2025},
  publisher={Elsevier}
}

@article{cerrato2024deep,
  title={A deep learning driven algorithmic pipeline for autonomous navigation in row-based crops},
  author={Cerrato, Simone and Mazzia, Vittorio and Salvetti, Francesco and Martini, Mauro and Angarano, Simone and Navone, Alessandro and Chiaberge, Marcello},
  journal={IEEE Access},
  year={2024},
  publisher={IEEE}
}

@incollection{martini2023enhancing,
  title={Enhancing navigation benchmarking and perception data generation for row-based crops in simulation},
  author={Martini, Mauro and Eirale, Andrea and Tuberga, Brenno and Ambrosio, Marco and Ostuni, Andrea and Messina, Francesco and Mazzara, Luigi and Chiaberge, Marcello},
  booktitle={Precision agriculture'23},
  pages={451--457},
  year={2023},
  publisher={Wageningen Academic}
}

@inproceedings{salvetti2023waypoint,
  title={Waypoint Generation in Row-Based Crops with Deep Learning and Contrastive Clustering},
  author={Salvetti, Francesco and Angarano, Simone and Martini, Mauro and Cerrato, Simone and Chiaberge, Marcello},
  booktitle={Machine Learning and Knowledge Discovery in Databases: European Conference, ECML PKDD 2022, Grenoble, France, September 19--23, 2022, Proceedings, Part VI},
  pages={203--218},
  year={2023},
  organization={Springer}
}

@article{winterhalter2021localization,
author = {Winterhalter, Wera and Fleckenstein, Freya and Dornhege, Christian and Burgard, Wolfram},
title = {Localization for precision navigation in agricultural fields—Beyond crop row following},
journal = {Journal of Field Robotics},
volume = {38},
number = {3},
pages = {429-451},
year = {2021}
}

@inproceedings{SLAMAgri,
  author={Khan, Md Sakif Uddin and Ahmed, Tanvir and Chen, Heping},
  booktitle={2025 IEEE 15th International Conference on CYBER Technology in Automation, Control, and Intelligent Systems (CYBER)}, 
  title={A Review of SLAM Techniques for Agricultural Robotics: Challenges, Architectures, and Future Directions}, 
  year={2025},
  volume={},
  number={},
  pages={598-603},
  doi={10.1109/CYBER67662.2025.11168347}}

@INPROCEEDINGS{AgriRobot,
  author={Singh, Shivendra and Vaishnav, Ram and Gautam, Saurabh and Banerjee, Somnath},
  booktitle={2024 2nd International Conference on Artificial Intelligence and Machine Learning Applications Theme: Healthcare and Internet of Things (AIMLA)}, 
  title={Agricultural Robotics: A Comprehensive Review of Applications, Challenges and Future Prospects}, 
  year={2024},
  volume={},
  number={},
  pages={1-8},
  doi={10.1109/AIMLA59606.2024.10531517}}

@article{devInAgri,
author = {Ding, Haizhou and Zhang, Baohua and Zhou, Jun and Yan, Yaxuan and Tian, Guangzhao and Gu, Baoxing},
title = {Recent developments and applications of simultaneous localization and mapping in agriculture},
journal = {Journal of Field Robotics},
volume = {39},
number = {6},
pages = {956-983},
doi = {https://doi.org/10.1002/rob.22077},
url = {https://onlinelibrary.wiley.com/doi/abs/10.1002/rob.22077},
eprint = {https://onlinelibrary.wiley.com/doi/pdf/10.1002/rob.22077},
year = {2022}
}

@INPROCEEDINGS{3DMoveToSee,
  author={Lehnert, Chris and Tsai, Dorian and Eriksson, Anders and McCool, Chris},
  booktitle={2019 IEEE/RSJ International Conference on Intelligent Robots and Systems (IROS)}, 
  title={3D Move to See: Multi-perspective visual servoing towards the next best view within unstructured and occluded environments}, 
  year={2019},
  volume={},
  number={},
  pages={3890-3897},
  doi={10.1109/IROS40897.2019.8967918}}

@Article{RevCanopy,
AUTHOR = {Wang, Yunfei and Jia, Weidong and Ou, Mingxiong and Wang, Xuejun and Dong, Xiang},
TITLE = {A Review of Orchard Canopy Perception Technologies for Variable-Rate Spraying},
JOURNAL = {Sensors},
VOLUME = {25},
YEAR = {2025},
NUMBER = {16},
ARTICLE-NUMBER = {4898},
URL = {https://www.mdpi.com/1424-8220/25/16/4898},
PubMedID = {40871760},
ISSN = {1424-8220},
DOI = {10.3390/s25164898}
}

@INPROCEEDINGS{ChangingOrchards,
  author={Pan, Siyu and Hu, Yaohua and Ohya, Akihisa and Yorozu, Ayanori},
  booktitle={2025 13th International Conference on Control, Mechatronics and Automation (ICCMA)}, 
  title={Robust Localization for Agricultural Robots in Seasonally Changing Orchards Using Edge-Based LiDAR Segmentation and IMU Fusion}, 
  year={2025},
  volume={},
  number={},
  pages={135-141},
  doi={10.1109/ICCMA67641.2025.11369692}}

@misc{desilva2025,
      title={Semantic-Aware Particle Filter for Reliable Vineyard Robot Localisation}, 
      author={Rajitha de Silva and Jonathan Cox and James R. Heselden and Marija Popovic and Cesar Cadena and Riccardo Polvara},
      year={2025},
      eprint={2509.18342},
      archivePrefix={arXiv},
      primaryClass={cs.RO},
      url={https://arxiv.org/abs/2509.18342}, 
}

@ARTICLE{agriPlane,
    
AUTHOR={Aguiar, André Silva  and Neves dos Santos, Filipe  and Sobreira, Héber  and Boaventura-Cunha, José  and Sousa, Armando Jorge },
           
TITLE={Localization and Mapping on Agriculture Based on Point-Feature Extraction and Semiplanes Segmentation From 3D LiDAR Data},
          
JOURNAL={Frontiers in Robotics and AI},
          
VOLUME={Volume 9 - 2022},
  
YEAR={2022},
  
URL={https://www.frontiersin.org/journals/robotics-and-ai/articles/10.3389/frobt.2022.832165},
  
DOI={10.3389/frobt.2022.832165},
  
ISSN={2296-9144}}

@Article{Epstein2017,
author={Epstein, Russell A.
and Patai, Eva Zita
and Julian, Joshua B.
and Spiers, Hugo J.},
title={The cognitive map in humans: spatial navigation and beyond},
journal={Nature Neuroscience},
year={2017},
month={Nov},
day={01},
volume={20},
number={11},
pages={1504-1513},
issn={1546-1726},
doi={10.1038/nn.4656},
url={https://doi.org/10.1038/nn.4656}
}

@inproceedings{yolo,
  title = {You {{Only Look Once}}: {{Unified}}, {{Real-Time Object Detection}}},
  shorttitle = {You {{Only Look Once}}},
  booktitle = {2016 {{IEEE Conference}} on {{Computer Vision}} and {{Pattern Recognition}} ({{CVPR}})},
  author = {Redmon, Joseph and Divvala, Santosh and Girshick, Ross and Farhadi, Ali},
  date = {2016-06},
  pages = {779--788},
  publisher = {{IEEE}},
  location = {{Las Vegas, NV, USA}},
  doi = {10.1109/CVPR.2016.91},
  url = {http://ieeexplore.ieee.org/document/7780460/},
  urldate = {2023-11-23},
  eventtitle = {2016 {{IEEE Conference}} on {{Computer Vision}} and {{Pattern Recognition}} ({{CVPR}})},
  isbn = {978-1-4673-8851-1},
  langid = {english}
}

@ARTICLE{Moreno2020-bf,
  title     = "On-ground vineyard reconstruction using a {LiDAR-based}
               automated system",
  author    = "Moreno, Hugo and Valero, Constantino and Bengochea-Guevara,
               Jos{\'e} Mar{\'\i}a and Ribeiro, {\'A}ngela and Garrido-Izard,
               Miguel and And{\'u}jar, Dionisio",
  journal   = "Sensors (Basel)",
  publisher = "MDPI AG",
  volume    =  20,
  number    =  4,
  pages     = "1102",
  month     =  feb,
  year      =  2020,
  copyright = "https://creativecommons.org/licenses/by/4.0/",
  language  = "en"
}

@article{JIANG2024108870,
title = {Navigation system for orchard spraying robot based on 3D LiDAR SLAM with NDT\_ICP point cloud registration},
journal = {Computers and Electronics in Agriculture},
volume = {220},
pages = {108870},
year = {2024},
issn = {0168-1699},
doi = {https://doi.org/10.1016/j.compag.2024.108870},
url = {https://www.sciencedirect.com/science/article/pii/S0168169924002618},
author = {Saike Jiang and Peng Qi and Leng Han and Limin Liu and Yangfan Li and Zhan Huang and Yajia Liu and Xiongkui He},
}

@misc{botsort, 
      title={BoT-SORT: Robust Associations Multi-Pedestrian Tracking}, 
      author={Nir Aharon and Roy Orfaig and Ben-Zion Bobrovsky},
      year={2022},
      eprint={2206.14651},
      archivePrefix={arXiv},
      primaryClass={cs.CV},
      url={https://arxiv.org/abs/2206.14651}, 
}

@article{aguiar2020localization,
  title={Localization and mapping for robots in agriculture and forestry: A survey},
  author={Aguiar, Andr{\'e} Silva and Dos Santos, Filipe Neves and Cunha, Jos{\'e} Boaventura and Sobreira, H{\'e}ber and Sousa, Armando Jorge},
  journal={Robotics},
  volume={9},
  number={4},
  pages={97},
  year={2020},
  publisher={MDPI}
}

@article{zhai2020decision,
  title={Decision support systems for agriculture 4.0: Survey and challenges},
  author={Zhai, Zhaoyu and Mart{\'\i}nez, Jos{\'e} Fern{\'a}n and Beltran, Victoria and Mart{\'\i}nez, N{\'e}stor Lucas},
  journal={Computers and Electronics in Agriculture},
  volume={170},
  pages={105256},
  year={2020},
  publisher={Elsevier}
}

@article{Bigelow:263079,
      author = {Bigelow, Daniel  and Borchers, Allison },
      journal = {Economic Information Bulletin Number 178},
      title = {Major Uses of Land in the United States, 2012},
      number = {1476-2017-4340},
      recid = {263079},
      pages = {69},
      address = {2017-08-28},
      year = {2017},
}

@article{diao2025localization,
  title={Localization technologies for smart agriculture and precision farming: A review},
  author={Diao, Zhihua and Chen, Lele and Yang, Yuanyuan and Liu, Yuchen and Yan, Jingyi and He, Shengxian and Zhang, Baohua},
  journal={Computers and Electronics in Agriculture},
  volume={236},
  pages={110464},
  year={2025},
  publisher={Elsevier}
}

@article{wu2025review,
  title={Review on key technologies for autonomous navigation in field agricultural machinery},
  author={Wu, Hongxuan and Wang, Xinzhong and Chen, Xuegeng and Zhang, Yafei and Zhang, Yaowen},
  journal={Agriculture},
  volume={15},
  number={12},
  pages={1297},
  year={2025},
  publisher={MDPI}
}

@article{droukas2023survey,
  title={A survey of robotic harvesting systems and enabling technologies},
  author={Droukas, Leonidas and Doulgeri, Zoe and Tsakiridis, Nikolaos L and Triantafyllou, Dimitra and Kleitsiotis, Ioannis and Mariolis, Ioannis and Giakoumis, Dimitrios and Tzovaras, Dimitrios and Kateris, Dimitrios and Bochtis, Dionysis},
  journal={Journal of Intelligent \& Robotic Systems},
  volume={107},
  number={2},
  pages={21},
  year={2023},
  publisher={Springer}
}

@article{hua2025harvesting,
  title={Key technologies in apple harvesting robot for standardized orchards: A comprehensive review of innovations, challenges, and future directions},
  author={Hua, Wanjia and Zhang, Zhao and Zhang, Wenqiang and Liu, Xiaohang and Hu, Can and He, Yichuan and Mhamed, Mustafa and Li, Xiaolong and Dong, Haoxuan and Saha, Chayan Kumer and others},
  journal={Computers and Electronics in Agriculture},
  volume={235},
  pages={110343},
  year={2025},
  publisher={Elsevier}
}

@article{feng2020yield,
  title={Yield estimation in cotton using UAV-based multi-sensor imagery},
  author={Feng, Aijing and Zhou, Jianfeng and Vories, Earl D and Sudduth, Kenneth A and Zhang, Meina},
  journal={Biosystems Engineering},
  volume={193},
  pages={101--114},
  year={2020},
  publisher={Elsevier}
}

@ARTICLE{IMU_pre,
  author={Forster, Christian and Carlone, Luca and Dellaert, Frank and Scaramuzza, Davide},
  journal={IEEE Transactions on Robotics}, 
  title={On-Manifold Preintegration for Real-Time Visual--Inertial Odometry}, 
  year={2017},
  volume={33},
  number={1},
  pages={1-21},
  doi={10.1109/TRO.2016.2597321}}

@INPROCEEDINGS{isam2,
  author={Kaess, Michael and Johannsson, Hordur and Roberts, Richard and Ila, Viorela and Leonard, John and Dellaert, Frank},
  booktitle={2011 IEEE International Conference on Robotics and Automation}, 
  title={iSAM2: Incremental smoothing and mapping with fluid relinearization and incremental variable reordering}, 
  year={2011},
  volume={},
  number={},
  pages={3281-3288},
  doi={10.1109/ICRA.2011.5979641}}
